\documentclass{article}

\usepackage[round,semicolon]{natbib}

\usepackage[final]{neurips_data_2022}

\usepackage{titlesec}
\usepackage[utf8]{inputenc} %
\usepackage[T1]{fontenc}    %
\usepackage{hyperref}       %
\hypersetup{
    colorlinks=true,
	linkcolor=blue,
	filecolor=magenta,      
	urlcolor=blue,
	citecolor=black,
	pdfinfo={
		Title={Forecasting Future World Events with Neural Networks},
		Author={Andy Zou, Tristan Xiao, Ryan Jia, Joe Kwon, Mantas Mazeika, Richard Li, Dawn Song, Jacob Steinhardt, Owain Evans, Dan Hendrycks},
		Subject={Forecasting, ML Safety, AI X-Risk},
		Keywords={forecast, forecasting, prediction market, judgmental forecasting, superforecasting, ai forecast, metaculus, ai oracle, calibration, uncertainty, ml safety, x-risk, existential risk, ai safety}
	}
}
\usepackage{url}            %
\usepackage{booktabs}       %
\usepackage{amsfonts}       %
\usepackage{nicefrac}       %
\usepackage{microtype}      %
\usepackage{xcolor}         %

\usepackage{url}
\usepackage{lipsum}
\usepackage{bbm}
\usepackage{booktabs}
\usepackage{tabularx}
\usepackage{makecell}
\newcolumntype{Y}{>{\centering\arraybackslash}X}
\newcolumntype{s}{>{\hsize=.3\hsize}Y}
\newcolumntype{t}{>{\hsize=.7\hsize}X}
\newcolumntype{b}{X}
\newcolumntype{u}{>{\hsize=0.8\hsize}Y}
\usepackage{wrapfig}
\usepackage{graphicx}
\usepackage{subcaption}
\usepackage{multirow}
\usepackage{epsfig}
\usepackage{graphicx}
\usepackage{xcolor}
\usepackage{amsmath}
\usepackage{amssymb}
\usepackage{subcaption}
\usepackage{textcomp}
\usepackage{gensymb}
\usepackage{mathtools}
\usepackage{amssymb}
\usepackage{wrapfig}
\usepackage{lipsum}
\usepackage{diagbox}
\usepackage{stfloats}
\usepackage{multirow}
\usepackage{tabularx}
\newcolumntype{Y}{>{\centering\arraybackslash}X}
\usepackage{fancyvrb}
\usepackage{arydshln}
\usepackage{makecell}
\usepackage{dsfont}
\usepackage{booktabs}
\usepackage{wrapfig}
\usepackage{enumitem}
\usepackage[frozencache,cachedir=.]{minted}

\newcolumntype{?}{!{\vrule width 1.1pt}}

\usepackage{letltxmacro}
\makeatletter
\let\oldr@@t\r@@t
\def\r@@t#1#2{%
\setbox0=\hbox{$\oldr@@t#1{#2\,}$}\dimen0=\ht0
\advance\dimen0-0.2\ht0
\setbox2=\hbox{\vrule height\ht0 depth -\dimen0}%
{\box0\lower0.4pt\box2}}
\LetLtxMacro{\oldsqrt}{\sqrt}
\renewcommand*{\sqrt}[2][\ ]{\oldsqrt[#1]{#2}}
\makeatother

\usepackage{algorithm}
\usepackage{algcompatible}
\algnewcommand\algorithmicreturn{\textbf{return}}
\algnewcommand\RETURN{\State \algorithmicreturn}%
\algnewcommand\algorithmicfunction{\textbf{function}}
\algdef{SE}[FUNCTION]{Function}{EndFunction}%
   [2]{\algorithmicfunction\ \text{#1}\ifthenelse{\equal{#2}{}}{}{(#2)}}%
   {\algorithmicend\ \algorithmicfunction}%

\title{Forecasting Future World Events \\ with Neural Networks}

\author{%
  Andy Zou \\
  UC Berkeley
  \And
  Tristan Xiao \\
  UC Berkeley
  \And
  Ryan Jia \\
  UC Berkeley
  \And
  Joe Kwon \\
  MIT
  \AND
  Mantas Mazeika \\
  UIUC
  \And
  Richard Li \\
  UC Berkeley
  \And
  Dawn Song \\
  UC Berkeley
  \And
  Jacob Steinhardt \\
  UC Berkeley
  \AND
  Owain Evans \\
  University of Oxford
  \And
  Dan Hendrycks \\
  UC Berkeley
}

\begin{document}

\maketitle

\begin{abstract}
Forecasting future world events is a challenging but valuable task. Forecasts of climate, geopolitical conflict, pandemics and economic indicators help shape policy and decision making. In these domains, the judgment of expert humans contributes to the best forecasts. Given advances in language modeling, can these forecasts be automated? To this end, we introduce Autocast, a dataset containing thousands of forecasting questions and an accompanying news corpus. Questions are taken from forecasting tournaments, ensuring high quality, real-world importance, and diversity. The news corpus is organized by date, allowing us to precisely simulate the conditions under which humans made past forecasts (avoiding leakage from the future). Motivated by the difficulty of forecasting numbers across orders of magnitude (e.g. global cases of COVID-19 in 2022), we also curate IntervalQA, a dataset of numerical questions and metrics for calibration. We test language models on our forecasting task and find that performance is far below a human expert baseline. However, performance improves with increased model size and incorporation of relevant information from the news corpus. In sum, Autocast poses a novel challenge for large language models and improved performance could bring large practical benefits.\looseness=-1
\end{abstract}
\vspace{-10pt}

\section{Introduction}

Forecasting plays a crucial role in the modern world. Climate forecasts shape the policies of governments and companies \citep{gillingham2018modeling}. Economic forecasts influence investment and employment \citep{christensen2018uncertainty}. In 2020, forecasts about the spread of COVID-19 led to national lockdowns and border closures \citep{adam2020modelling}, slowing the spread of the virus. Consequently, machine learning (ML) models that make accurate forecasts across a broad range of topics could enable more informed decision making at scale and improve ML safety \citep{hendrycksmlsafety2021}.

Two main approaches to forecasting are described in the forecasting literature: statistical and judgmental forecasting \citep{webby1996judgemental, armstrong2001principles}. In \textit{statistical forecasting}, forecasts are made by traditional statistical models for time-series prediction such as autoregression \citep{makridakis2008forecasting} or by ML time-series models \citep{makridakis2020forecasting,triebe2021neuralprophet}. Humans create and tune the models but do not tweak individual forecasts. This works well when there are many past observations of the variable being forecast and minimal distribution shift. By contrast, in \textit{judgmental forecasting} human forecasters use their own judgment to determine forecasts. The forecasters may use statistical models, but often integrate information from various sources including news, accumulated knowledge, and \textit{a priori} reasoning. This enables forecasting for questions where past data is scarce or subject to distribution shift \citep{tetlock2016superforecasting}. For brevity, we refer to judgmental forecasting as ``forecasting'' in the rest of the paper.

Because it relies on scarce human expertise, forecasting is only used for a small number of questions. This motivates using ML to automate forecasting, e.g.\ by automating human information retrieval (finding news sources), reasoning (to decide if some evidence bears on a forecast), and quantitative modeling. ML models may also have some advantages over human forecasters. Models can read through text or data much faster than humans and can discern patterns in noisy high-dimensional data that elude humans. When it comes to learning, humans cannot be trained on past data in manner simulating actual forecasting (e.g.\ How likely was the Soviet Union’s collapse from the viewpoint of 1980?) because they know the outcomes -- but past data can be used for ML models.

As a step towards automating human forecasting, we introduce \textit{Autocast}, a new dataset for measuring ML models' forecasting ability. Autocast includes thousands of forecasting questions collected from human forecasting tournaments. The questions vary in the forecasting horizon from days to decades, in the topic (including politics, economics and science), and in the answer format (e.g.\ multiple-choice vs.\ predicting a number). The questions are pre-selected for public interest, and there is a strong human baseline (the crowd aggregate of many competitive forecasters). The questions in Autocast are about past events (e.g.\ the US 2020 election) and so ML models could answer them simply by memorizing what happened. To test forecasting ability, we need to simulate the state of information \textit{before} the past events (``retrodiction’’). To this end, we curate a corpus of news items from Common Crawl \citep{ccnews} that is organized by date. This means a model can be exposed only to news from before the outcomes being forecast, allowing for a rigorous test of retrodiction.\looseness=-1

\begin{figure}[t]
    \vspace{-10pt}
    \centering
    \includegraphics[width=\textwidth]{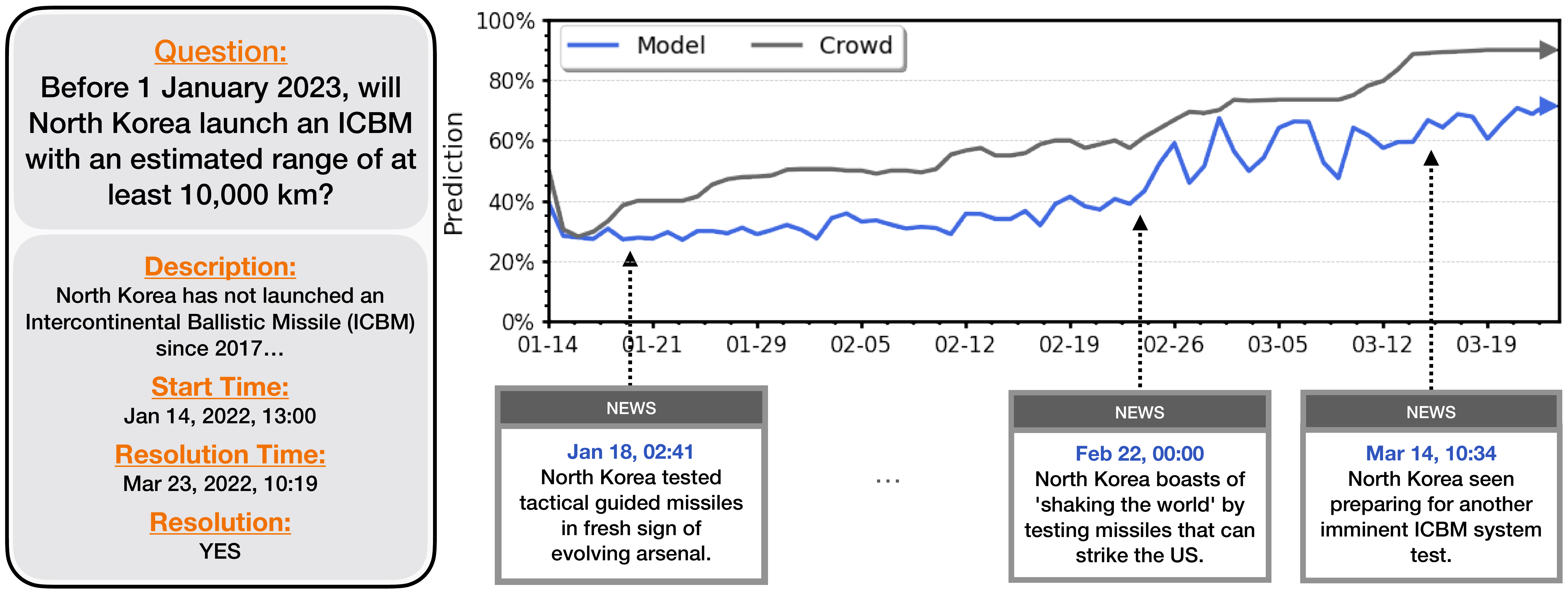}
    \vspace{5pt}
    \caption{Example from the Autocast dataset, including the question, the resolution of the question, and the timeseries of aggregate human expert forecasts (Crowd) from the start date to the time the question resolves. We train a language model to generate forecasts at each timestep, using only news articles available at that timestep (i.e.\ without allowing any leakage of information from the future).}
    \label{fig:banner}
    \vspace{15pt}
\end{figure}

We implement a number of baseline models on Autocast, and demonstrate how language models can be trained on past forecasting questions by retrieving from our news corpus. We find that performance improves with model size and that information retrieval helps. However, all baselines are substantially worse than aggregate human forecasts. On forecasting binary outcomes, the best ML model achieves 65\% accuracy vs.\ 92\% for humans (and 50\% for random). The same ML model \citep{2020t5} is close to the human ceiling when fine-tuned on other NLP benchmarks (e.g.\ SQuAD from \citet{rajpurkar2016squad}), which shows that Autocast is a challenging, real-world test for ML. Experiment code and the dataset are available at {\color{blue} \href{ https://github.com/andyzoujm/autocast}{github.com/andyzoujm/autocast}}.

\paragraph{Contributions.}
\begin{enumerate}[leftmargin=*]
    \item 
We introduce Autocast, a dataset for forecasting that covers diverse topics (e.g.\ politics, economics, society, science) and varying time horizons.
    \item 
Part of our dataset is a large news corpus organized by date, allowing us to rigorously evaluate model performance on historical forecasts.
    \item
We show that forecasting is challenging for current language models, with accuracy and calibration far below a strong human baseline.
\end{enumerate}

\begin{table}[t]
\vspace{-10pt}
\begin{center}
{
\setlength\tabcolsep{3pt}
\setlength\extrarowheight{3pt}
\begin{tabular}{cccc}\\
Question Summary & Category & Answer Type &  Resolution \\\toprule
\makecell{Will a Tesla car demonstrate fully \\autonomous capability before the end of 2021?} & Science \& Tech & T/F &  No \\  \midrule
\makecell{What will be Putin's approval rating value 3 \\months after the potential invasion of Ukraine?}  & Politics & Numerical & 83\\  \midrule
\makecell{When will the US-Canada border reopen?} & Social & Numerical & Nov 8, 2021\\ \midrule
\makecell{How many vacancies will arise on the U.S. Supreme \\
Court in 2021? (A) 0 (B) 1 (C) 2 (D) 3 or more} & Economy & MCQ & A \\\bottomrule
\end{tabular}
}
\vspace{10pt}
\caption{Examples from the Autocast dataset. For brevity, we do not depict the full question specification, which often includes context, definitions, and detailed resolution criteria.}
\label{tab:dataset_examples}
\end{center}
\end{table}

\section{Related Work}

\begin{wrapfigure}{r}{0.5\textwidth}
    \vspace{-5pt}
    \centering
    \includegraphics[width=0.48\textwidth]{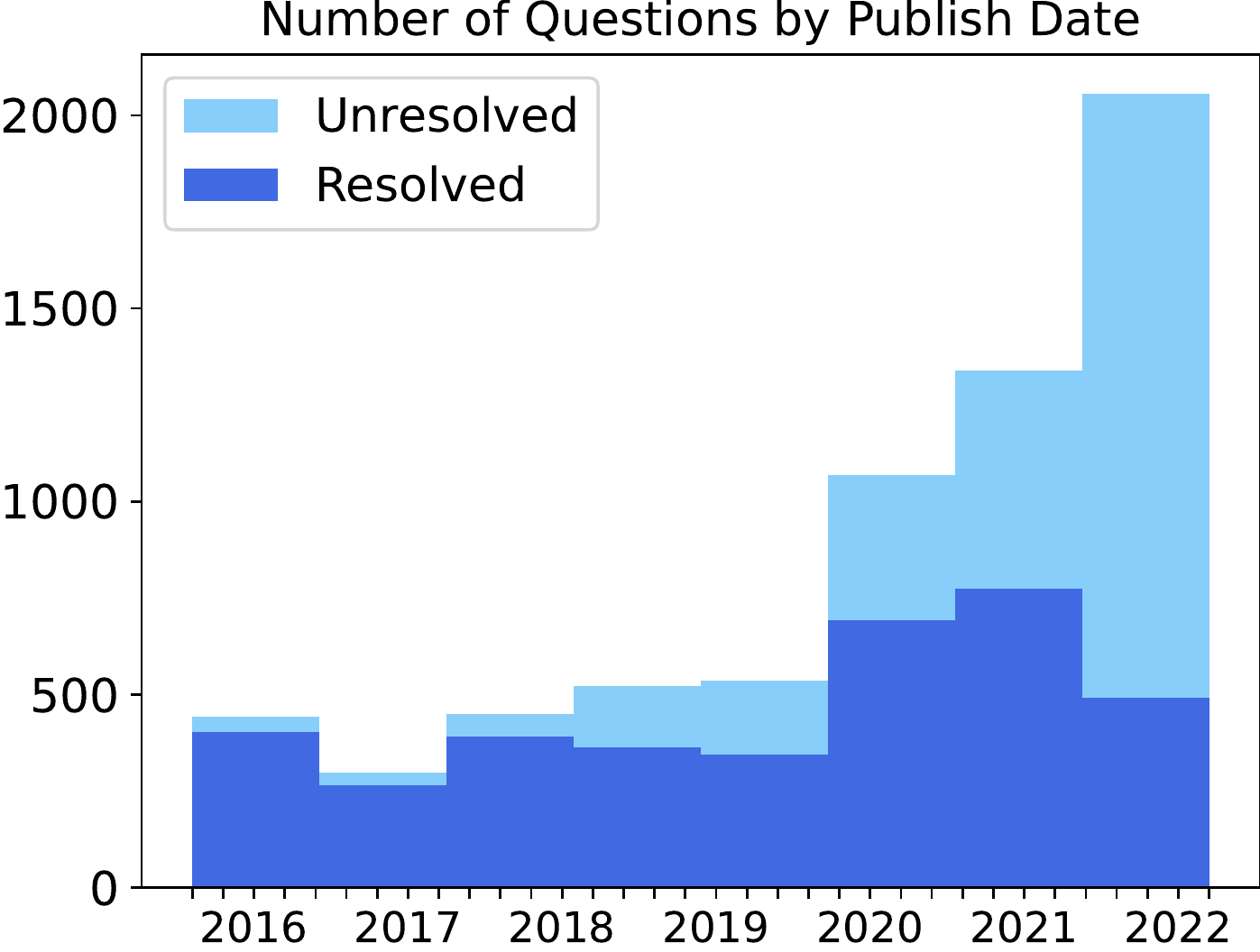}
    \caption{The number of questions in Autocast by publish date. Unresolved questions are about events after 2022 (e.g.\ the 2024 US Election). They are not included in the test set but can be used as auxiliary training data. Note that the number of questions is accelerating. Future questions will be added to Autocast, improving it over time.}
    \label{fig:question_counts}
    \vspace{-5pt}
\end{wrapfigure}

\textbf{Forecasting.}\quad
A recent experiment \citep{mathias} tested GPT-3 in the few-shot setting on true/false questions collected from Metaculus (one of the sources for Autocast). However, since questions were not filtered by date, some answers would have appeared in GPT-3's training data. Similar to our work, ForecastQA \citep{jin2020forecastqa} is a dataset of forecasting questions that covers a range of topics. However, ForecastQA's questions were written by crowdworkers without forecasting experience. Consequently, the questions are often nonsensical or ambiguous given the lack of additional context, e.g. ``To how many people will the Representative of an internet speak to by September 2019?'', or ``In July 2019, will an article say there were no volunteers in 2016?''. We found that a high percentage of ForecastQA questions suffer from these issues. By contrast, our questions were written by experienced forecasters and are always unambiguous given the full question description. Finally, ForecastQA’s human baseline was done retrospectively (making it unrealistic) whereas our dataset contains expert human forecasts from real forecasting questions.

\textbf{Information Retrieval.}\quad
Information retrieval is crucial for forecasting, as good forecasts depend on up-to-date, specialized information drawn from multiple sources \citep{tetlock2016superforecasting}. Recent work has used information retrieval to improve question-answering in large language models \citep{lewis2020retrieval,nakano2021webgpt,shuster2021retrieval} or to address time-sensitive questions \citep{chen2021a}. This has been applied to tasks that are related to forecasting, such as fact checking and truthful question-answering. In forecasting, it is useful to read and compare multiple news articles daily, in order to build an accurate picture of the current state, and then to iterate this process. We design an architecture for this purpose (albeit with limits on article length and time horizon), drawing inspiration from \citet{wang2020fly}.

\textbf{Calibration.}\quad
Calibration is important in forecasting \citep{tetlock2016superforecasting}. Even expert forecasters will be highly uncertain about some outcomes of interest. Such forecasts will be more useful in the form of calibrated probabilities than as point estimates. Thus forecasters are evaluated with proper scoring rules, which incentivize calibration. There is an extensive literature on improving the calibration of deep learning models \citep{guo2017calibration,nguyen2015posterior,lin2022teaching,minderer2021revisiting,kull2019beyond}, mostly for classification with a fixed set of classes. One part of Autocast requires models to forecast continuous quantities varying over multiple orders of magnitude, which has not been explored in prior work.

\textbf{Truthful question-answering.}\quad
Current language models often generate falsehoods when answering questions \citep{shuster2021retrieval,lin2021truthfulqa}, and they also achieve poor calibration when giving probabilistic answers \citep{hendryckstest2021} to human knowledge questions. However, for questions with a known ground truth answer, we expect models to improve as a result of scale, fine-tuning, and information-retrieval from reliable sources \citep{bai2022training,nakano2021webgpt,HadfieldMenell2016CooperativeIR,NEURIPS2020_f50a6c02,wainwright2019safelife}. Yet humans also want models to give calibrated and truthful answers to questions that are too difficult or costly for us to answer ourselves \citep{irving2018ai, evans2021truthful,Leike2017AISG,unsolved,reddy2020learning,nahian2021training}. Forecasting is useful for this purpose. Forecasting questions are challenging but eventually become easy to evaluate. By contrast, it may be difficult for humans to evaluate superior answers to open problems in fundamental philosophy or science.  %

\section{The Autocast Dataset}

\begin{figure}[t]
    \vspace{-10pt}
    \centering
    \includegraphics[width=\textwidth]{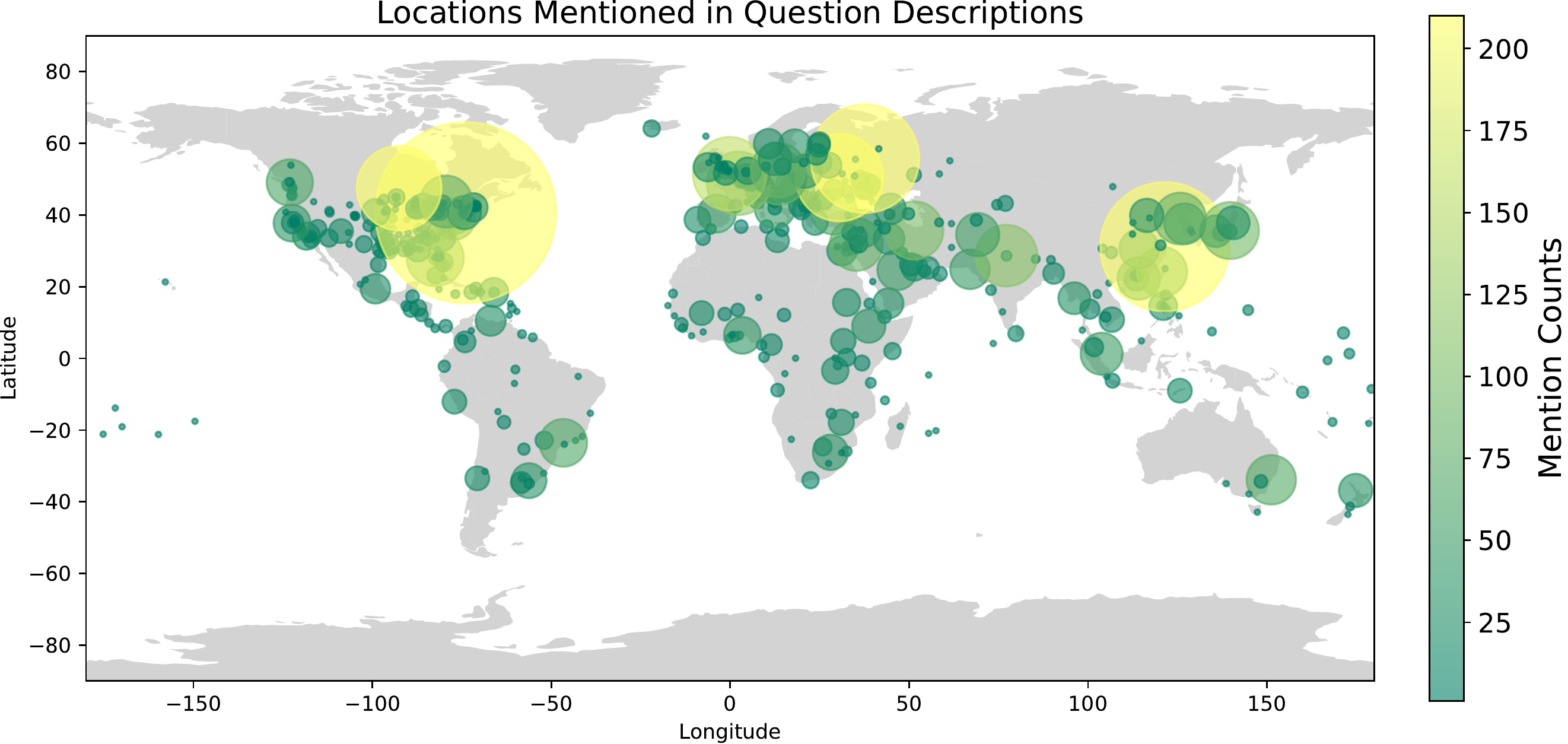}
    \caption{Autocast contains questions about locations across the world. The questions in the dataset mention over $500$ cities, spanning six continents.}
    \label{fig:geographic}
    \vspace{5pt}
\end{figure}

\noindent\textbf{Forecasting Questions.}\quad We collected all available forecasting questions from three public forecasting tournaments (Metaculus, Good Judgment Open, and CSET Foretell), which resulted in 6,707 questions total. These questions tend to have broad public interest (e.g., national rather than local elections) and clear resolution criteria. Most questions are not already covered well by specialized forecasts (such as weather forecasts). The questions are either true/false, multiple-choice, or involve forecasting a numerical quantity or date (see Table~\ref{tab:dataset_examples} for examples). In these forecasting tournaments, participants begin forecasting a question on a given day (the ``start date'') and update their forecasts multiple times up until the ``close date.'' At some later time, the forecast is \textit{resolved} and participants are scored based on all their forecasts. (Note the resolution date is often just after the closing date but not always. The resolution can also happen \textit{before} the planned closing date: e.g.\ when forecasting when an event will occur.) Thus the ``crowd’’ forecast (which aggregates over participants) is a time-series of forecasts from the start to close date.

Autocast includes the question, the start and close dates, the answer (if the question has resolved), and the time-series of crowd forecasts (Figure~\ref{fig:banner}). Half of the questions have not yet resolved and correspond to ongoing tournaments. Some of these questions concern events decades in the future, requiring reasoning over long time horizons. These questions can still be used as training data by using the crowd forecast as the target (as a high-quality proxy for the ground truth). However, the test set only includes resolved questions. Our dataset also includes metadata that is helpful for forecasting. There is detailed background information about the question (including precise terms of resolution) and also links to relevant information posted by tournament participants. We include more details in the appendix.

\begin{figure}[t]
    \centering
    \includegraphics[width=\textwidth]{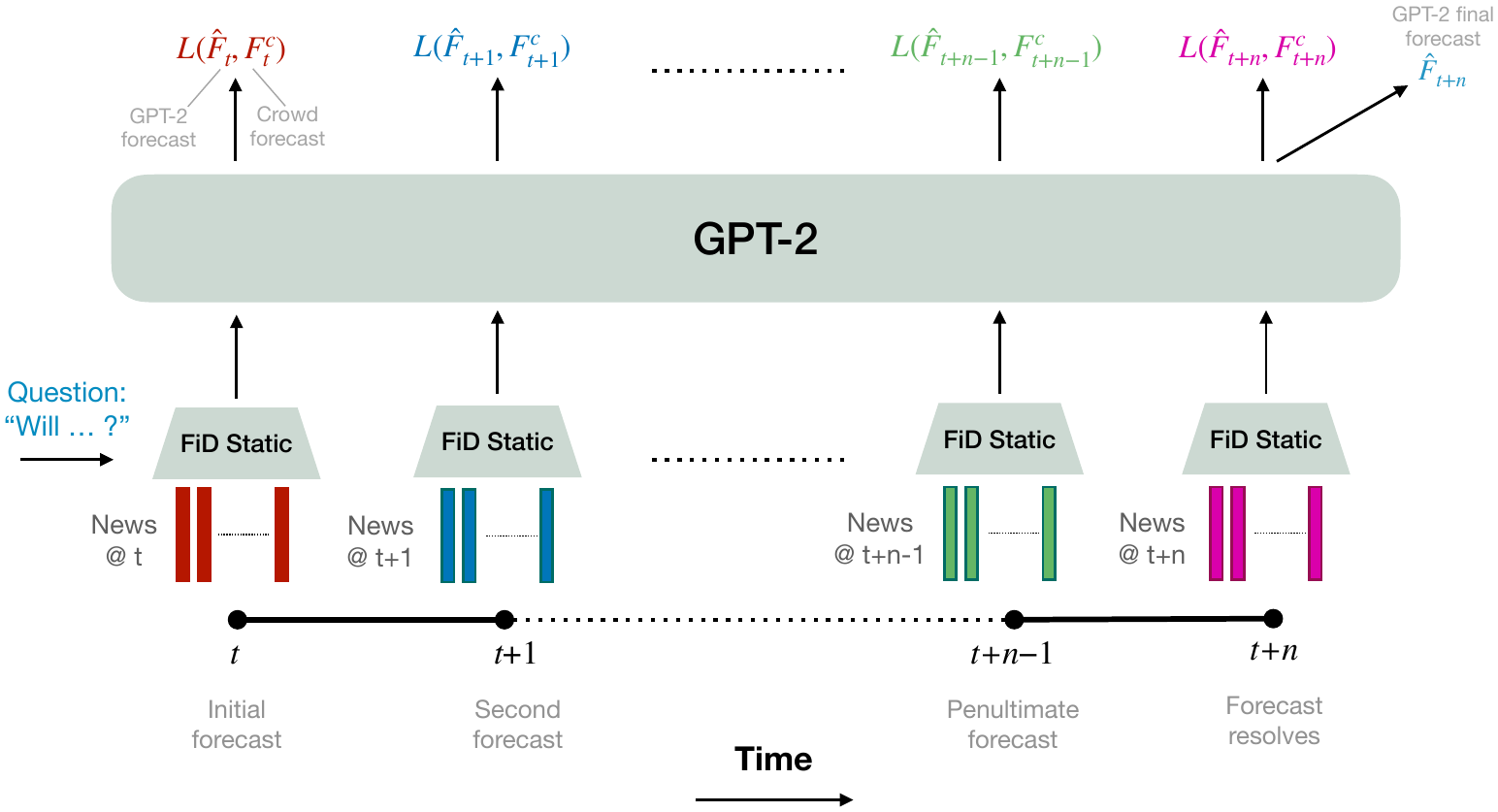}
    \vspace{5pt}
    \caption{\textbf{Illustration of our FiD Temporal model.} Forecasts are made each day (from start date to resolution) by GPT-2. The input to GPT-2 is the top-1 daily news article retrieved by BM25, which is encoded by FiD Static (a T5 model). In training, GPT-2's target is the average of daily crowd predictions (denoted `$F_t^c$' for day $t$) and the resolved outcome. Like human forecasters, GPT-2 accumulates news information over time and updates its predictions. 
    }
    \label{fig:method}
    \vspace{5pt}
\end{figure}

\noindent\textbf{Train and test split.}\quad
It is standard in ML for the test set to be drawn from the same distribution as the train set. However, randomly splitting our questions into train and test without considering the date would not simulate the conditions of forecasting. For example, a test question (``Will Trump win the 2020 election?’’) could come from an earlier date than a related training question (``Will President Biden pass the stimulus?’’). Thus, we split our questions using a date cut-off of mid-2021, which means that questions in the test set resolve from mid-2021 to mid-2022. Note that if a model is pre-trained on data from after mid-2021, this will also not simulate forecasting faithfully. In both train and test sets, we implement dataset balancing for the true/false questions. To flip a label, we negate the question using OpenAI’s GPT-3-175B Edit model \citep{brown2020language} and manually check for correct negation.

\noindent\textbf{Contemporaneous news as context for forecasts.}\quad When a human is making a forecast at time $t$, they use past and present ($\leq t$) information sources but are not exposed to any information from the future ($>t$). If they forecast again at $t+1$, they will have updated on new information that was generated from $t$ to $t+1$. These conditions can be simulated for ML models by (a) pre-training on text generated before time $t$, and (b) providing the model with new information generated between $t$ and $t+1$. To this end, we provide a corpus of news articles scraped from CommonCrawl news \citep{ccnews, Hamborg2017} that is organized by publish date. The articles were derived from diverse sources between 2016 to mid-2022 and total more than 200GB of data.

\subsection{Dataset Analysis}

\noindent\textbf{Distribution of Questions.}\quad
The questions in Autocast cover a very wide variety of topics. We divide the questions into five main categories: Economy, Politics, Science, Social, and Other. Each category contains numerous subcategories for a total of $44$ subcategories ranging from foreign policy to AI. We list all subcategories in the Supplementary Material. The questions also cover a wide geographical distribution, as shown in Figure~\ref{fig:geographic}. Overall, Autocast tests both breadth of subject matter and depth (since questions ask for quantitative predictions about a specific, operationalized variable).

\noindent\textbf{Adding new questions over time.}\quad
The number of questions submitted to forecasting platforms is rapidly increasing (Figure~\ref{fig:question_counts}). If trends continue, in two years there will be twice as many questions available. Autocast is a living dataset and will be updated periodically with new questions. This will provide both more data for training and a new set of test questions (to assess overfitting). 

\noindent\textbf{Human forecasts.}\quad
The human crowd forecast for a given question becomes more accurate from the start to closing date, as shown in Figure \ref{fig:performance_over_time}. This is what we would expect if humans are updating their forecasts as more information comes out. In contrast to most ML benchmarks, the human crowd judgments are probabilistic. This allows us to evaluate their calibration. In the Supplementary Material, we show that crowd forecasts are well-calibrated.

\noindent\textbf{Distribution shift.}\quad
We expect a distribution shift over time in both the questions being asked and in the answers. For example, there will be fewer questions about Ukraine before 2022. This distribution shift is inherent to forecasting and so it is crucial that models can manage it. We do find a shift in the distribution of question categories. For example, the number of questions in the Social category increased from 12.6\% in the training set to 28.2\% in the test set, possibly due the Covid-19 pandemic (which is included in Social). %

\begin{table}[t]
\begin{center}
{
\setlength\tabcolsep{10pt}
\setlength\extrarowheight{2pt}
\begin{tabular}{l c | c c c | c ? c}
Model & Parameters & T/F & MCQ & Numerical & Score & Average \\
\hline
Random &-- &50.0 &22.1 &34.5 &18.8 &18.8 \\
\hline
\multirow{3}{7em}{UnifiedQA}
&0.2B &45.4 &23.5 &34.5 &17.2 & \multirow{3}{*}{19.5} \\ %
&0.8B &48.2 &23.5 &34.5 &18.6\\ %
&2.8B &54.9 &25.1 &34.5 &22.8 \\
\hline
\multirow{3}{7em}{T5}
&0.2B &61.3 &24.0 &20.5 &32.4 & \multirow{3}{*}{32.9} \\ %
&0.8B &60.0 &29.1 &21.7 &33.7 \\ %
&2.8B &60.0 &26.8 &21.9 &32.5 \\
\hline
\multirow{3}{7em}{FiD Static}
&0.2B &62.0 &29.6 &24.5 &33.5 & \multirow{3}{*}{37.2} \\ %
&0.8B &64.1 &32.4 &21.8 &37.4 \\ %
&2.8B &\textbf{65.4} &35.8 &19.9 &\textbf{40.6} \\
\hline
\multirow{3}{7em}{FiD Temporal} 
& 0.6B &62.0 &33.5 &23.9 &35.8 & \multirow{3}{*}{\textbf{37.8}} \\ %
&1.5B &63.8 &32.4 &21.0 &37.6 \\ 
&4.3B &62.9 &\textbf{36.9} &\textbf{19.5} & 40.1 \\
\hline
\textcolor{gray}{Human Crowd} & \textcolor{gray}{--} & \textcolor{gray}{92.4} & \textcolor{gray}{81.0} & \textcolor{gray}{8.5} & \textcolor{gray}{82.5} & \textcolor{gray}{82.5} \\
\Xhline{2.5\arrayrulewidth}
\end{tabular}
}
\vspace{20pt}
\caption{Model accuracy on the Autocast dataset for each question type: true/false (T/F), multiple-choice question (MCQ), and numerical (Numerical). For Numerical, lower is better. For other metrics, higher is better. The model FiD Static (based on T5) retrieves the top 10 news articles over the period, while FiD Temporal (based on GPT-2 with T5 encoder) retrieves the top 1 article each day. Averaging over all model sizes, we find that the FiD Temporal achieves the best average. %
}
\vspace{-10pt}
\label{tab:big_results}
\end{center}
\end{table}

\section{Experiments}

\subsection{Baselines}

The \textit{Crowd} baseline uses the final aggregate human forecast before the closing date. The \textit{Random} baseline uses the analytically computed random accuracy for true/false and multiple-choice questions. For numerical questions, random predictions are sampled uniformly from the bounded range of possible answers specified in each question.

\textbf{Models without retrieval.}\quad
We evaluate \textit{UnifiedQA-v2} \citep{khashabi2022unifiedqa} and \textit{T5} \citep{2020t5} models of various sizes. These models are trained on a variety of tasks, enabling strong generalization on many unseen language tasks. Using zero-shot prompting for UnifiedQA, we report results on classification questions. The UnifiedQA models were not trained on numerical questions, hence, we report random performance to enable comparison with other baselines. T5 is fine-tuned using its original output head for true/false and multiple-choice questions. To output numerical answers with T5, we add an additional linear output head.

\begin{wrapfigure}{r}{0.5\textwidth}
    \centering
    \includegraphics[width=0.5\textwidth]{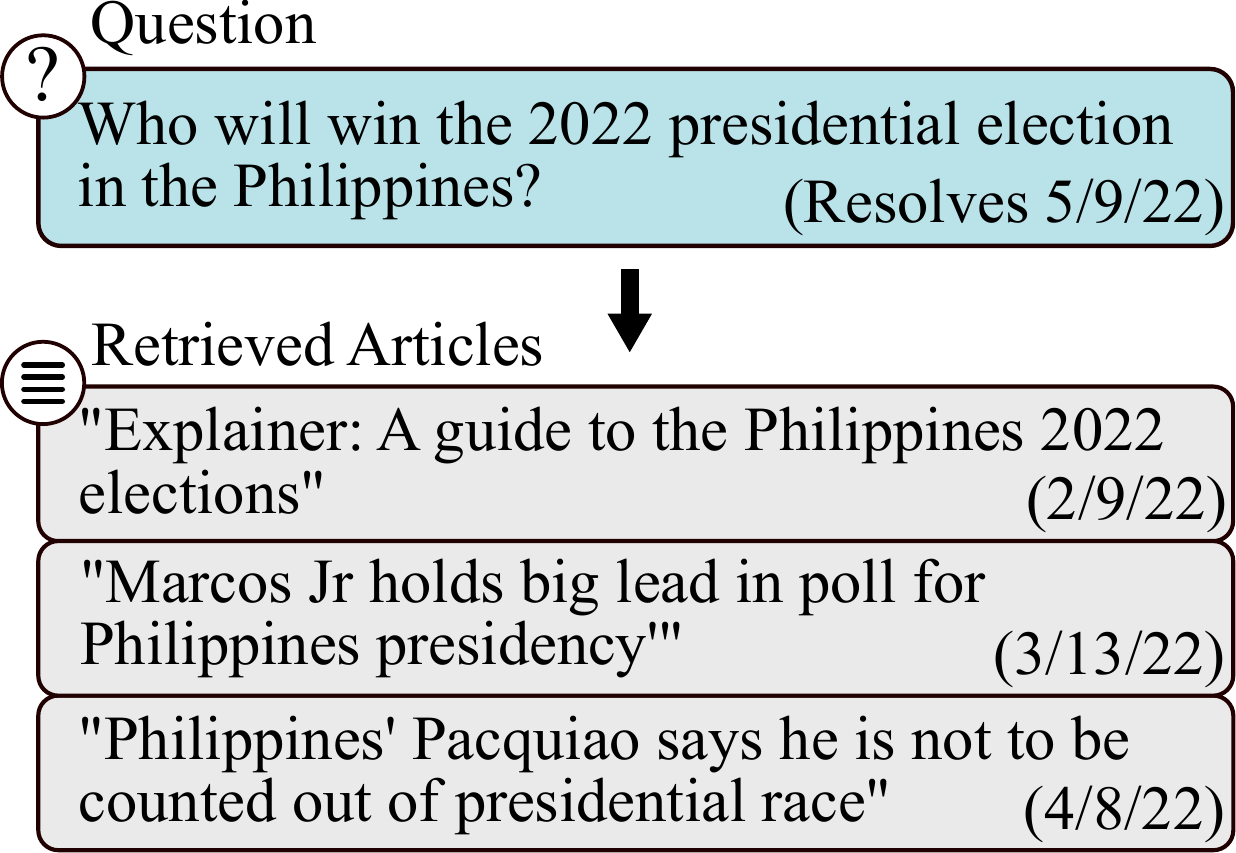}
    \vspace{5pt}
    \caption{Articles retrieved by BM25 for a Politics question in the Autocast dataset with publication dates in parentheses. The articles are retrieved from 200GB of news and are highly relevant to making an informed forecast.}
    \label{fig:retrieval_examples}
\end{wrapfigure}

\textbf{Retrieval-Based Methods.}\quad
We investigate whether retrieval models can improve performance by selecting relevant articles from the news corpus included with Autocast. Importantly, news articles after the close time or resolution time of a question are not available for retrieval, so retrieved articles only include information about the past. For all retrieval methods, we use Fusion-in-Decoder or FiD \citep{Izacard2021LeveragingPR} to encode articles retrieved by BM25 \citep{Robertson1994OkapiAT, thakur2021beir} with cross-encoder reranking. FiD uses T5 to encode retrieved passages along with the question and can be viewed as a minimal extension of T5 for incorporating retrieval. We truncate retrieved articles to a maximum length of $512$ tokens.

The \textit{FiD Static} baseline uses the top $10$ retrieved articles after reranking, which is the standard method for retrieval-augmented prediction. The \textit{FiD Temporal} baseline leverages the intermediate crowd predictions (before the question is resolved) as auxiliary supervision. The intuition is that crowd predictions will change based on rational incorporation of new evidence, and these updates will not be captured by just training on the final outcome.
For each day between the question's open and close date, we generate an embedding of the top news article using the frozen fine-tuned FiD Static model.
These embeddings are then treated as input embeddings to an autoregressive model (GPT-2 \citep{radford2019language}), which is fine-tuned to predict the average of the daily crowd prediction and the ground truth outcome. We illustrate this method in Figure \ref{fig:method}. Figure \ref{fig:banner} shows predictions from an FiD Temporal model over time for an example question.

\subsection{Metrics}
For true/false and multiple-choice questions, we evaluate models using percent accuracy. For numerical questions, we use $\ell_1$ distance, bounded between $0\%$ and $100\%$. We denote these question types as T/F, MCQ, and Numerical, respectively. To evaluate aggregate performance, we use a combined Score metric $(\text{T/F} + \text{MCQ} - \text{Numerical})/2$, which has an upper bound of $100\%$. A score of $100\%$ indicates perfect prediction on all three question types. Note that since numerical question responses are normalized between $0\%$ and $100\%$, the combined Score metric is lower-bounded at $-50\%$. We also report the Average score, which averages the combined metric of all model sizes.

\subsection{Forecasting Evaluation}
\textbf{Setup.}\quad
We fine-tune the T5 baseline for $10$ epochs with a batch size of $8$, an initial learning rate of $5\times 10^{-5}$ with linear decay schedule, and a weight decay of $1\times 10^{-2}$. The maximum sequence length of the T5 model is set to 512. We train FiD Temporal models for $5$ epochs with a constant learning rate of $5\times 10^{-5}$. Hyperparameters are selected based on early experiments. Additional details are in the Supplementary Material.

\textbf{Results.}\quad
We show results in Table \ref{tab:big_results}. Although UnifiedQA-v2 obtains strong performance on various natural language benchmarks, it obtains close to random zero-shot performance on Autocast, showing the difficulty of forecasting. Fine-tuned T5 performs better, but multiple-choice accuracy is still at nearly random chance levels. Retrieval-based methods substantially outperform both UnifiedQA-v2 and T5, showing a relative increase in the Average score of $93\%$ and $15\%$, respectively. Moreover, retrieval-based methods become more effective as parameter count increases, which suggests that the models learn to extract relevant information from retrieved articles.

Comparing the FiD Static and FiD Temporal baselines, we see that the Average score is slightly higher for FiD Temporal. However, the largest FiD Static model has the highest individual score. Thus, our temporal training strategy for incorporating the auxiliary crowd predictions neither harms nor helps compared to the static retrieval baseline. Future work could develop more effective ways of using these auxiliary training signals.

\subsection{Model Analysis}

\begin{wrapfigure}{r}{0.5\textwidth}
    \centering
    \includegraphics[width=0.48\textwidth]{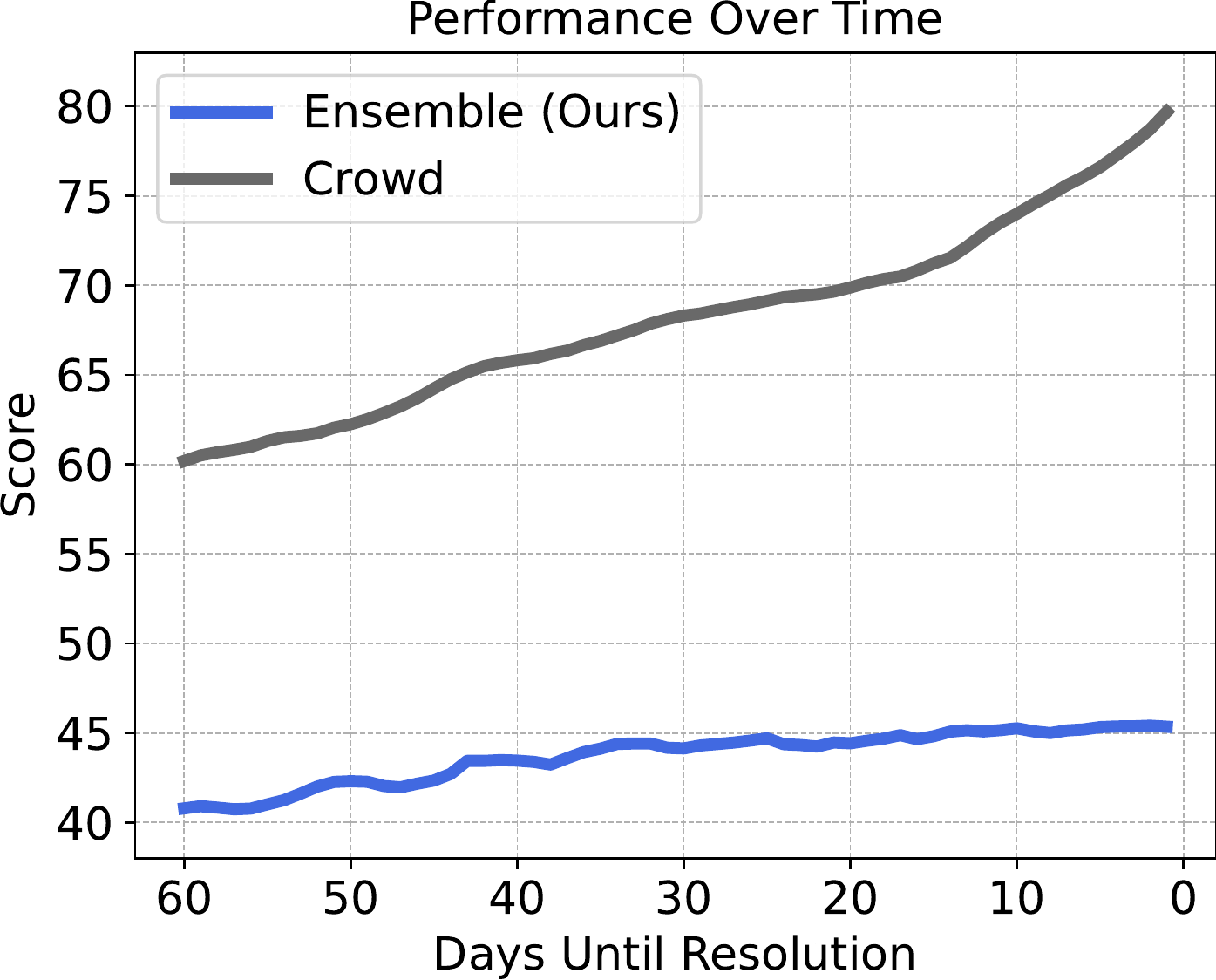}
    \vspace{5pt}
    \caption{For the crowd and an ensemble of the two largest FiD Temporal models, prediction score increases as the resolution date grows nearer. %
    This trend may be due to more relevant information becoming available over time, which the model can access through retrieval from the news corpus.}
    \label{fig:performance_over_time}
\end{wrapfigure}

\textbf{Relevance of Retrieved Articles.}\quad We find that the retrieved articles are often highly relevant to the question. In Figure \ref{fig:retrieval_examples}, we show examples of articles retrieved by BM25 from the news corpus in Autocast. Baseline models have access to the article text, but for brevity we only show the article title. The articles give information that is clearly relevant for making an informed forecast. Note that the T5 backbone for the baselines was pre-trained on data from before 2020, far before the timeline of the question, so retrieval provides vital information that models would not otherwise have. This suggests that large improvements on Autocast could come from integrating information from retrieved articles more effectively. We expect that more sophisticated retrieval methods would also improve performance, although efficiency becomes a concern when using large retrieval methods.

\textbf{Detailed Performance.}\quad
In the Supplementary Material, we show the performance of baseline methods on a more granular level. The per-category results indicate that Science \& Technology is the most challenging category for models, whereas human forecasters have relatively consistent performance across categories. Inspecting the subset of questions that have been active for at least two months, we also find that the accuracy of the human crowd forecast and a model ensemble steadily increases over time up to the resolution date (Figure~\ref{fig:performance_over_time}). This is to be expected, as more information about the eventual outcome is available closer to the time (e.g.\ election polls become more accurate). For this plot, we show an ensemble of the two largest FiD Temporal models, which has slightly higher final performance than the individual models and a clearer trend over time.

\section{Calibrated Prediction of Numerical Quantities}
\label{sec:calibration}

In our results, we evaluate baselines on the accuracy of their point estimates, rather than their calibration. However, the eventual goal for Autocast is for models to achieve good calibration as well as accuracy. Here we describe an auxiliary dataset that helps with this goal for the challenging case of calibration on numerical questions.

\noindent\textbf{The IntervalQA Dataset.}\quad
In the Autocast training set, numerical quantities range over many orders of magnitude. Furthermore, Autocast has fewer than $1,\!000$ numerical training questions. This problem of making \textit{calibrated} predictions for quantities over many orders of magnitude using text inputs has not been addressed in work on calibration for language models. To this end, we curate IntervalQA, an auxiliary dataset of numerical estimation problems and provide metrics to measure calibration. The problems in the dataset are not forecasting problems but instead involve giving calibrated predictions for fixed numerical quantities. The questions were sourced from NLP datasets covering diverse topics and with answers varying across orders of magnitude: SQuAD, 80K Hours Calibration \citep{80khours}, Eighth Grade Arithmetic \citep{cobbe2021gsm8k}, TriviaQA \citep{DBLP:journals/corr/JoshiCWZ17}, Jeopardy, MATH \citep{hendrycks2021measuring}, and MMLU \citep{hendryckstest2021}. We filtered these datasets for questions with numerical answers, which yielded about 30,000 questions.

\subsection{Metrics}

We evaluate whether confidence intervals are calibrated. Concretely, if a method outputs $80\%$ confidence intervals on each test example, we would like the true prediction target to fall inside of these intervals $80\%$ of the time. Additionally, we would like for models to be calibrated across their entire dynamic range of outputs. To measure this, we compute \textit{RMS Calibration Error} similarly to \citet{nguyen2015posterior} and \citet{hendrycks2019oe}, but with fixed confidence levels $c \in \{50\%, 55\%, \dotsc, 95\%\}$ and such that calibration is sensitive to dynamic range. We describe this metric in detail in the Supplementary Material. Low RMS Calibration Error indicates that models are calibrated across their entire dynamic range. We also compute the median prediction error between the predicted point estimate and the ground-truth target (\textit{Point Estimate Distance}) and the median interval length averaged across all confidence levels (\textit{Conf. Interval Length}).

\subsection{Experiments}
We fine-tune DeBERTa-v3 models \citep{DBLP:journals/corr/abs-2006-03654} to predict a point estimate and a set of confidence intervals corresponding to the confidence levels in the RMS calibration error metric. On a high level, we use a loss with three components: (1) MSE loss between the predicted point estimate and the ground-truth target, (2) MSE loss between the boundaries of the predicted confidence intervals and the ground-truth target for boundaries that are on the wrong side of the target, (3) a penalty on the length of the predicted intervals to encourage finer predictions. The models are trained for $5$ epochs with a batch size of $100$. A detailed description is in the Supplementary Material. We show results in Table \ref{tab:conf_results}. All three metrics decrease with model size.

\begin{table}[t]
\begin{center}
{
\setlength\tabcolsep{10pt}
\setlength\extrarowheight{2pt}
\begin{tabular}{l | c c c}
Parameters & Point Estimate Distance & Conf. Interval Length & RMS Calibration Error \\
\hline
22M &20.8 &2072.4 &19.1 \\
44M &20.3 &1115.7 &16.6 \\
86M &19.6 &763.1 &16.9 \\
304M &\textbf{18.1} &\textbf{305.4} &\textbf{13.5} \\
\hline
\end{tabular}
}
\vspace{15pt}
\caption{Results for DeBERTa-v3 models trained to output confidence intervals on our dataset of numerical predictions. The high dynamic range of the targets leads to large confidence intervals, but median interval size decreases with larger models as does RMS Calibration Error.}
\vspace{-10pt}
\label{tab:conf_results}
\end{center}
\end{table}

\section{Conclusion}

We introduced Autocast, a dataset for measuring the ability of neural networks to forecast future world events. The dataset contains thousands of forecasting questions from public forecasting tournaments, including ground truth outcomes and aggregated human predictions. We also curated a large corpus of news items from the Common Crawl news corpus, enabling rigorous evaluations without information leakage. We evaluated numerous baseline algorithms and demonstrated that model size and information retrieval can improve forecasting performance. To better evaluate calibration for numerical prediction, we introduced IntervalQA, a large collection of numerical prediction questions with a wide dynamic range of prediction targets, and evaluated state-of-the-art language models. Our results show significant room for future improvement.

\clearpage
\bibliography{main}
\bibliographystyle{plainnat}

\clearpage
\appendix
\section{Additional Experimental Details and Results}

\subsection{Autocast Experiments}
\textbf{Calibration Results.}\quad
In Figure \ref{fig:crowd_calib_perf_by_category}, we show the adaptive binning calibration curve for crowd forecasts on all resolved true/false questions by plotting the fraction of positives against the model's predicted probability for the positive class.

Additionally, we can compare the calibration of our static and temporal models to crowd performance on the resolved test set. Treating true/false questions as two-class classification problems and combining them with multiple-choice questions, we calculate adaptive binning calibration error with a bin size of 50 samples. The largest FiD Static model incurs a $40 \%$ calibration error while the human crowd incurs a much smaller $8 \%$ calibration error. By leveraging crowd predictions in our FiD Temporal models, we reduce the calibration error to $17 \%$, showing potential for improvements.

\textbf{Model and Training Loss.}\quad
The FiD Temporal model uses three separate linear heads after its hidden state outputs to answer each type of questions (true/false, multiple-choice, and numerical). In particular, the multiple-choice head has 12 outputs which is the maximum number of choices in the training set. Additionally, the original input embeddings are replaced with a linear layer to map from the FiD Static's hidden states to the GPT-2's hidden states. Finally, to make training more stable, we average the loss over the sequence of predictions for each question to weigh the questions evenly. Moreover, the losses of the three types of questions are normalized by their respective baseline loss (uniformly random predictions) before summing together so that their losses are on the same scale.

\textbf{Retrieval from CC-NEWS.}\quad
Given a question, for each day the question is active, we retrieve the top 10 relevant news articles from the daily articles. In our FiD-Temporal experiments, we only use the top 1 from every day. Then, we aggregate all these articles from different dates and rank them according to the retrieval score. The top 10 articles are used for the FiD-Static model. We follow the Terms of Use for the Common Crawl website. The dataset is fully reproducible with the script to download and filter CC-NEWS on GitHub.

\subsection{Confidence Intervals}

\begin{figure*}[ht]
    \begin{minted}[escapeinside=||,mathescape=true]{Python}
    
    Is = [0.5, 0.55, ..., 0.95]
    num_intervals = len(Is)
    
    def low_containment_mask(lowers, uppers, labels, Is):
        # lowers, uppers: Predicted lower and upper bounds of intervals
        # Is: Target confidence levels
        # Returns: A list of boolean values indicating which confidence level
        #          has containment ratio below the target level within batch
        contained = (lowers <= labels) * (labels <= uppers)
        ratio_contained = contained.mean(dim=0)
        return ratio_contained < Is
    
    def get_confidence_intervals(logits):
        # logits: Model output with (2 * num_intervals + 1) neurons
        deltas, point_estimates = |\color{blue}{softplus}|(logits[:, :-1]), logits[:, -1:]
        lower_deltas = deltas[:, :num_intervals]
        higher_deltas = deltas[:, num_intervals:]
        interval_lengths = lower_deltas + higher_deltas
        # custom cumsum with gradients accumulated once on each delta
        lower_deltas = utils.|\color{blue}{cumsum}|(lower_deltas)
        higher_deltas = utils.|\color{blue}{cumsum}|(higher_deltas)
        lowers = point_estimates - lower_deltas
        uppers = point_estimates + higher_deltas

        return lowers, uppers, point_estimates, interval_lengths
    
    out = |\color{blue}{get\_confidence\_intervals}|(logits)
    lowers, uppers, point_estimates, interval_lengths = out
    
    |$\mathcal{L}_p$| = |\color{blue}{MSE}|(point_estimates, labels)
    l_mask = lowers > labels
    u_mask = uppers < labels
    |$\mathcal{L}_b$| = |\color{blue}{MSE}|(lowers, labels) * l_mask + |\color{blue}{MSE}|(uppers, labels) * u_mask
    |$\mathcal{L}_i$| = |\color{blue}{MSE}|(interval_lengths, 0)
    # normalize loss by the label magnitude, adjusting for small labels
    |$\mathcal{L}_p$| /= (1 + |\color{blue}{abs}|(labels))
    |$\mathcal{L}_b$| /= (1 + |\color{blue}{abs}|(labels))
    |$\mathcal{L}_i$| /= (1 + |\color{blue}{abs}|(labels))
    # activate loss for particular confidence levels based on ci_mask
    ci_mask = |\color{blue}{low\_containment\_mask}|(lowers, uppers, labels, Is)
    |$\mathcal{L}_b$| = |$\mathcal{L}_b$|.mean(dim=0) * ci_mask
    |$\mathcal{L}_i$| = |$\mathcal{L}_i$|.mean(dim=0) * (1 - ci_mask)
    
    |$\alpha$|, |$\beta$|, |$\gamma$| = 1, 1, 0.01 # hyperparameters
    loss = |$\alpha$| * |$\mathcal{L}_p$|.mean() + |$\beta$| * |$\mathcal{L}_b$|.mean() + |$\gamma$| * |$\mathcal{L}_i$|.mean()
    
    \end{minted}
    \vspace{5pt}
    \caption{A reference implementation of the baseline training loss for outputting calibrated confidence intervals. For the confidence levels where too few true labels fall inside the predicted intervals, we encourage the model to adjust its boundaries through boundary loss $\mathcal{L}_b$. Conversely, we encourage the model to shrink the predicted intervals if too many fall inside the predicted intervals. \looseness=-1
    }\label{fig:pseudocode}
    \vspace{-10pt}
\end{figure*}

\textbf{Interval Construction.}\quad
In the reference implementation of the get\_confidence\_intervals function in Figure \ref{fig:pseudocode}, we construct our intervals by first producing a point estimate for each question and iteratively adding on non-negative, non-symmetric deltas on both sides, so that the intervals become nested and wider for higher confidence levels.

\begin{wraptable}{r}{0.5\textwidth}
\vspace{-15pt}
\begin{tabular}{l|ccc}\\ 
& Resolved & Unresolved & Total \\\midrule
Train & \makecell{2815 \\ \textcolor{gray}{4411}} & \makecell{1375 \\ \textcolor{gray}{1974}} & \makecell{4190 \\ \textcolor{gray}{6385}} \\  \midrule
Test & \makecell{907 \\ \textcolor{gray}{1292}} & \makecell{1610 \\ \textcolor{gray}{2305}} & \makecell{2517 \\ \textcolor{gray}{3597}} \\  \midrule
Total & \makecell{3722 \\ \textcolor{gray}{5703}} & \makecell{2985 \\ \textcolor{gray}{4279}} & \makecell{6707 \\ \textcolor{gray}{9982}} \\\bottomrule
\end{tabular}
\vspace{5pt}
\caption{The number of forecasting questions in Autocast. In total, there are nearly 10,000 questions. Gray text indicates the number of questions after augmenting true/false questions with their negations, a procedure we use to balance the dataset.}\label{table:data_splits}
\end{wraptable}

\textbf{Training Loss for Baseline.}\quad
First, because the labels span a large numerical range, we normalize them by taking the $log$. Then, we construct a loss with three components shown in Figure \ref{fig:pseudocode}: (1) $\mathcal{L}_p$: MSE loss between the predicted point estimate and the ground-truth target, (2) $\mathcal{L}_b$: MSE loss between the boundaries of the predicted confidence intervals and the ground-truth target for boundaries that are on the wrong side of the target, (3) $\mathcal{L}_i$: a penalty on the length of the predicted intervals to encourage finer predictions. Based on whether the ratio of true labels contained in the predicted intervals is higher than the target confidence level, we either activate the boundary loss $\mathcal{L}_b$ or the interval length loss $\mathcal{L}_i$ for that particular confidence level output. Lastly, the three components are weighted by coefficients $1,1,0.01$ chosen with a simple search using the validation set.

\textbf{Adaptive RMS Metric.}\quad An important task for numerical forecasting is outputting calibrated uncertainty estimates. However, a unique challenge in this setting is that answers can vary across many orders of magnitude. To evaluate the calibration of confidence intervals across a large dynamic range of output values, we design a specialized local calibration metric \citep{zhao2020individual, Kull2019BeyondTS}, shown in Algorithm \ref{algo:rms_calib_err}. First, test examples are sorted by their target value and split into bins with a fixed number of examples each (adaptive binning). Then, we calculate calibration error across all bins using a Euclidean norm \citep{hendrycks2019oe}. Finally, we average this local calibration error across all confidence levels, giving the final metric. For brevity, we refer to this overall metric as RMS Calibration Error. A low value for this error metric indicates that models are calibrated across their entire dynamic range of output values.

\begin{algorithm}[t]
	\begin{algorithmic}[1]
		\STATE \textbf{Input:} A set of $N$ examples each with label $\{y_i\}_{i=1}^N$ and $C$ predicted confidence intervals 
		$\{(l_i^c,u_i^c)\}_{c=1,i=1}^{C,N}$ corresponding to $C$ confidence levels $\{\mathcal{I}^c\}_{c=1}^C$ (e.g., $\mathcal{I}^C=0.95$). Set bin size to $M$.
	    \Function{AdaptiveRMS}{}
	    \STATE Sort the examples by labels $y_n$ in ascending order.
	    \STATE Assign a bin label $b_k = \left\lfloor\frac{k-1}{M}\right\rfloor+1$ to each by splitting sorted examples into chunks of $M$.
	    \STATE Let $\{B_i\}_{i=1}^b$ be the set of bins and $B_i$ the subset of examples in bin $i$.
		\FOR{$c = 1, \ldots, C$}
		\STATE Calculate empirical containment for bin $i$ $$\widehat{p}_i^c = \frac{1}{|B_i|} \sum_{k \in B_i} \mathds{1}(y_k \in [l^c_k, u^c_k])$$
		\STATE Calculate root mean squared calibration error $$\text{RMS}^c = \sqrt{\frac{1}{b} \sum_{i=1}^{b} (\widehat{p}_i^c - \mathcal{I}^c)^2}$$
		\ENDFOR
        \STATE Output $\frac{1}{C}\sum_{c=1}^C\text{RMS}^c$
        \EndFunction
	\end{algorithmic}
	\caption{\textsc{RMS} Calibration Error}\label{algo:rms_calib_err}
\end{algorithm}

\textbf{Calibration Dataset Statistics.}\quad
The dataset of numerical questions gathered for our calibration evaluations has training, validation, and test sets containing 32,200, 3,443, and 6,170 examples respectively.

\section{Additional Dataset Information}\label{appendix:dataset_info}

\subsection{Dataset Details}

\begin{figure}[t]
\centering
\begin{subfigure}{.44\textwidth}
  \centering
  \includegraphics[width=0.98\linewidth]{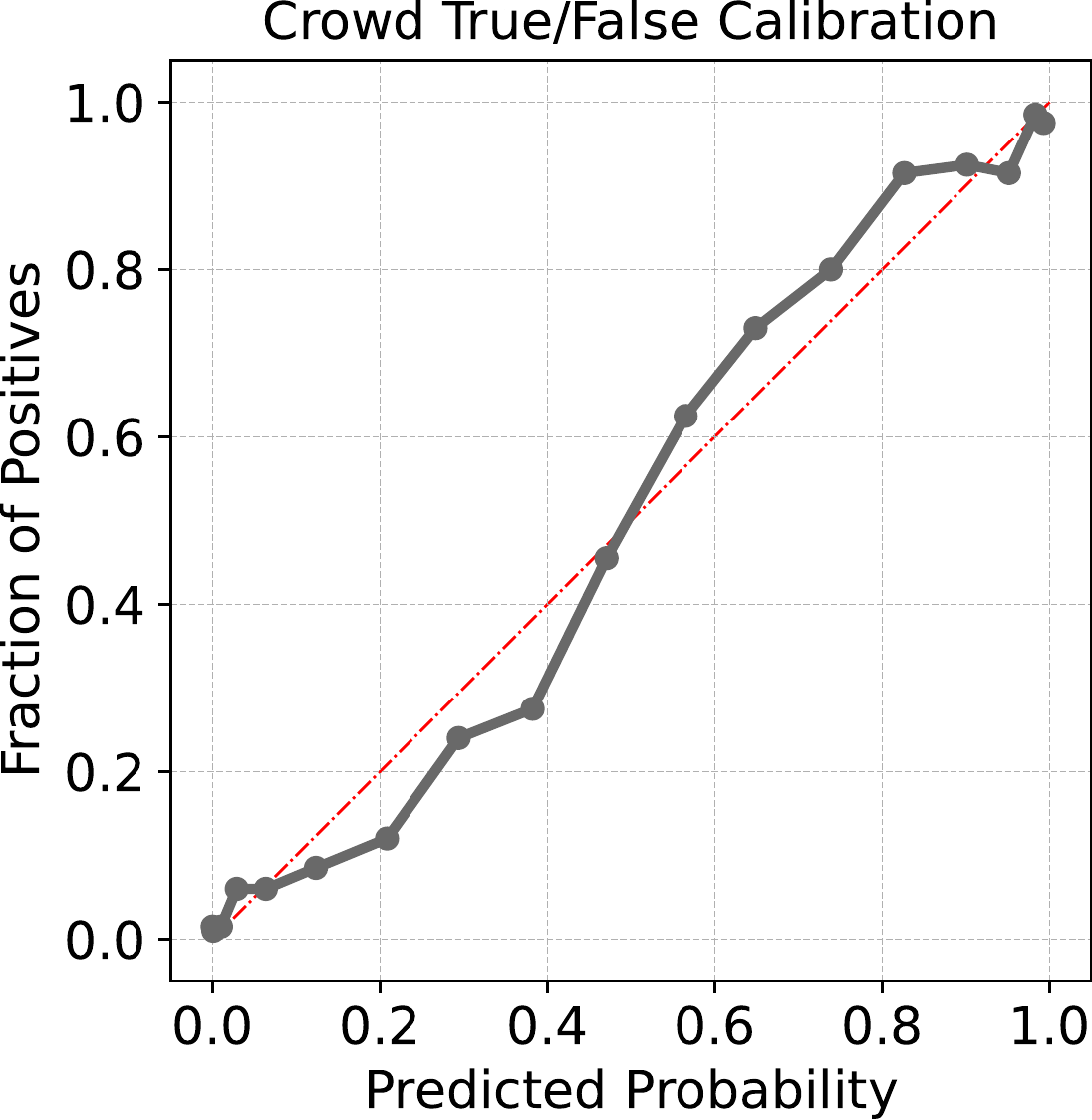}
  \label{fig:sub1}
\end{subfigure}%
\begin{subfigure}{.56\textwidth}
  \centering
  \includegraphics[width=0.98\linewidth]{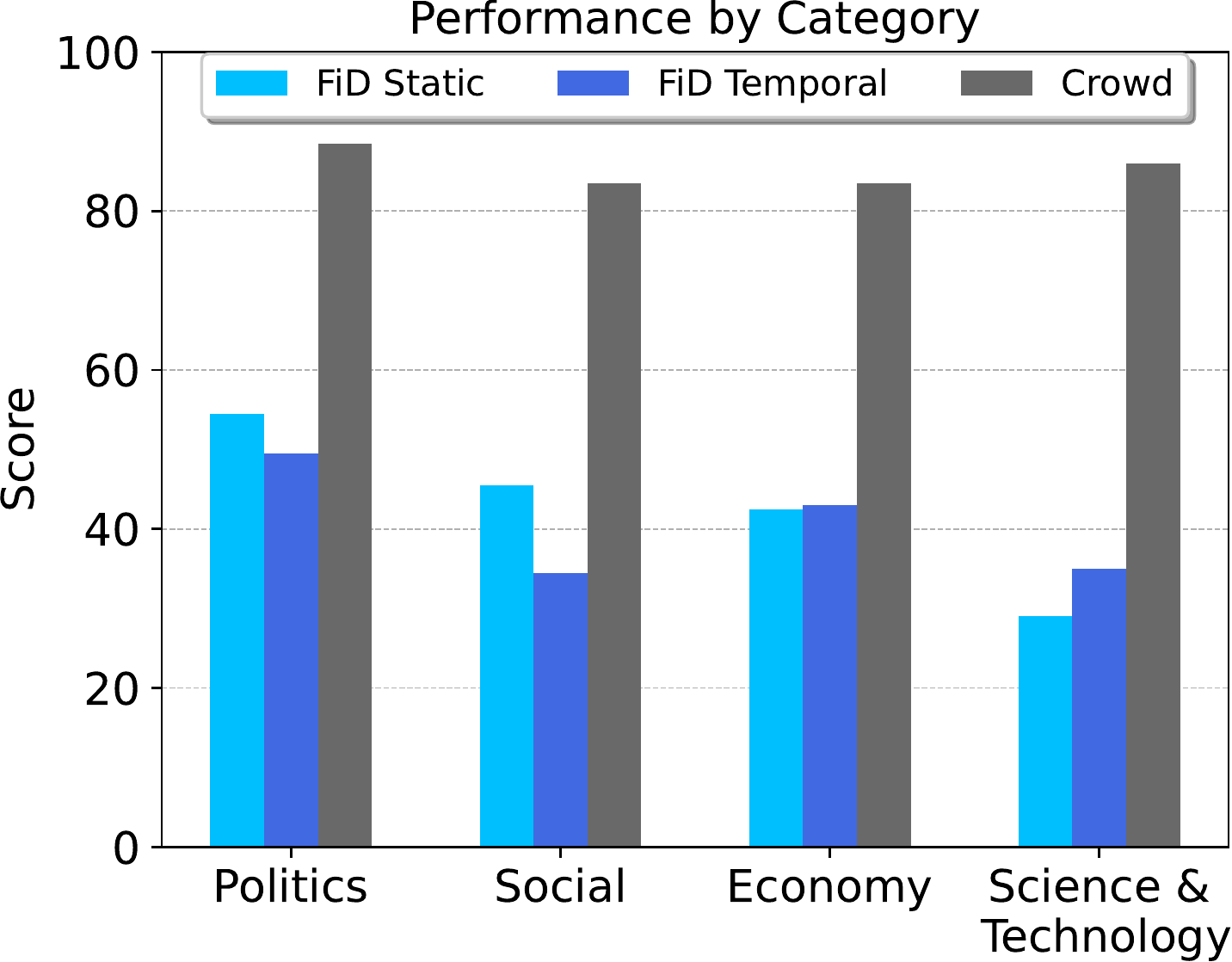}
  \label{fig:sub2}
\end{subfigure}
\caption{Left: Crowd forecasts for true/false questions have good calibration. Right: The per-category performance of baselines. Score indicates the combined score metric.}
\label{fig:crowd_calib_perf_by_category}
\end{figure}

\begin{wraptable}{r}{0.45\textwidth}
\vspace{-63pt}
\begin{tabular}{l|ccc}\\ 
& T/F & MCQ & Numerical \\\midrule
Train & 3187 & 753 & 471 \\  \midrule
Test & 775 & 176 & 341 \\  \midrule
Total & 3962 & 929 & 812 \\\bottomrule
\end{tabular}
\caption{The number of resolved questions in Autocast, grouped by question type.}\label{table:data_type_splits}
\vspace{-10pt}
\end{wraptable}

The Autocast dataset contains 6,707 unique questions in total, spanning three question types, including resolved and unresolved. After we balance the true/false questions by adding negated questions, the true/false question count doubles, making the grand total 9,757. The numbers of training and test examples are shown in Table \ref{table:data_splits} for ease of reference. The numbers below are based on the expanded dataset using true/false balancing. The Autocast training set we experiment with does not include unresolved questions. This training set contains 4,411 examples, and the test set contains 1,292 examples. To prevent leakage of future information, the train set consists of all questions that closed or resolved before 5-11-2021 and the test set consists of all questions that closed or resolved after 5-11-2021. In addition, we also release 1,974 unresolved train questions having a publish date before 5-11-2021 and 2,305 unresolved test questions published after 5-11-2021. Note that our baselines do not use any unresolved questions, so there is a guarantee of no leakage. However, training with auxiliary training signals from unresolved questions (e.g., crowd forecasts) requires additional care to ensure no leakage. Namely, crowd forecasts from after 5-11-2021 must not be used.

\begin{wrapfigure}{r}{0.5\textwidth}
    \centering
    \includegraphics[width=0.50\textwidth]{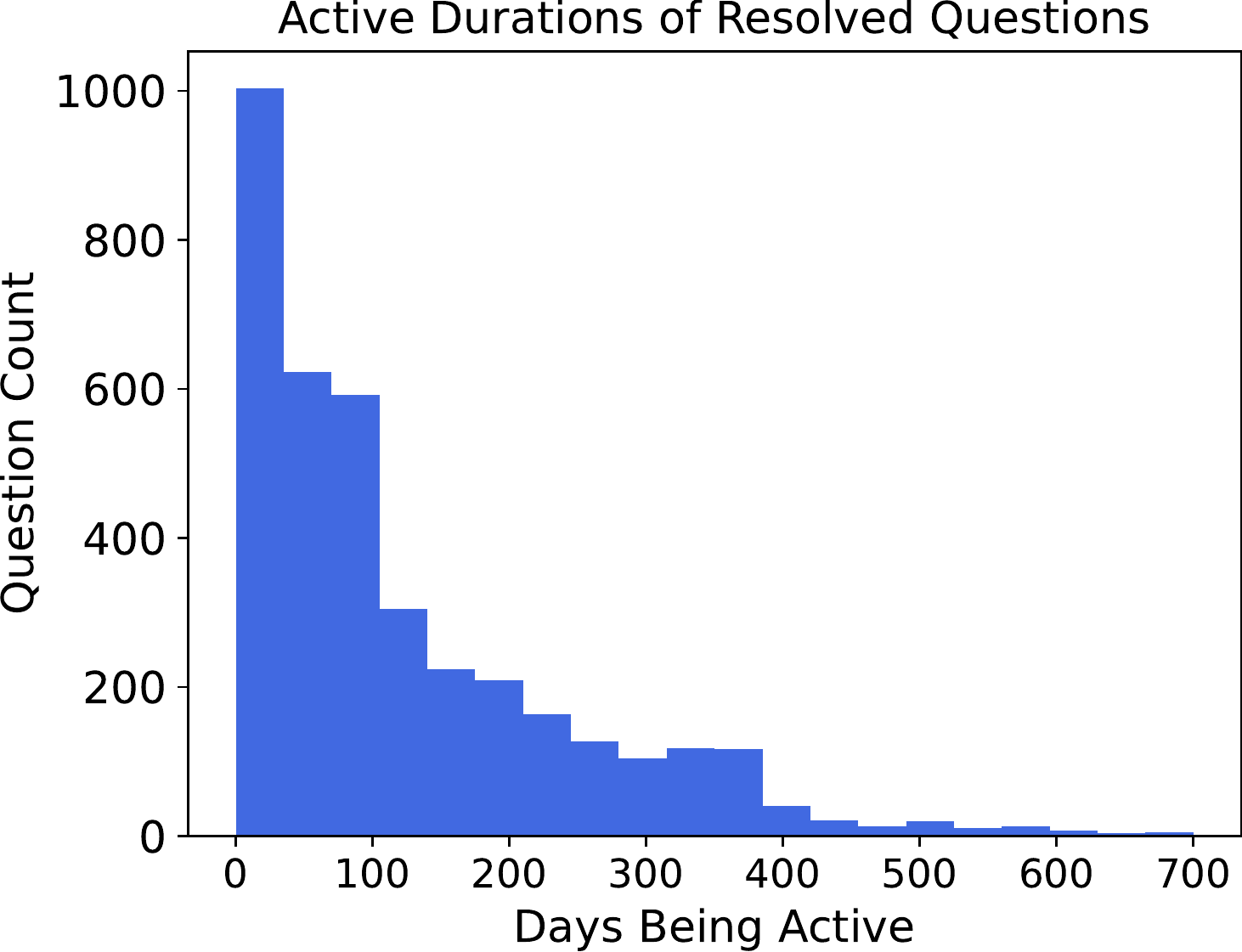}
    \caption{We visualize the distribution of the duration of the active periods for Autocast questions. Questions vary greatly in terms of how long they are active in the forecasting market, with questions taking up to years to resolve.}
    \label{fig:duration}
    \vspace{-20pt}
\end{wrapfigure}

\paragraph{Per-Category Performance.}
In Figure \ref{fig:crowd_calib_perf_by_category}, we show performance by category using the combined score metric. Science \& Technology questions are the most challenging for the FiD Static and FiD Temporal baselines, while the crowd predictions perform similarly on all question categories. There is a substantial gap between models and crowd predictions, but crowd predictions are still far from a perfect score of 100\%.

\paragraph{Computation of Crowd Forecasts.}
The human crowd forecasts are directly obtained from forecasting platforms, and the precise meaning depends on the platform. For example, for Metaculus questions the crowd forecast represents the median forecast with the recent player predictions weighted more. For Good Judgment Open questions, it represents the median of the recent 40\% of forecasts. In all cases, the crowd forecast aggregates previous individual forecasts at a given time.

\begin{table}[t]
\setlength\tabcolsep{15pt}
\begin{tabular}{ccc}\\ 
Category & Percentage & Subcategories \\\midrule
Politics & $31\%$ & \makecell{Geopolitics, Security and Conflict, Elections, \\Foreign Policy, Leader Entry/Exit, Law,\\ Economic Policy, US Policy, Ukraine} \\  \midrule
Social & $22\%$ & \makecell{COVID-19, Social Issues, Environment, \\ Effective Altruism, Sports, Entertainment, Health, \\ Society, Pandemic, Animal Welfare, \\ Metaculus, Climate, Education} \\ \midrule
Science \& Tech & $21\%$ & \makecell{Technology, Computing, Biological Sciences, \\Physical Sciences, Computer Science, Biology, \\ Human Sciences, AI, Mathematics, Tech} \\ \midrule
Economy & $20\%$ & \makecell{Business, Finance, Industry, Economic Indicators,\\ Infrastructure, Microelectronics, Semiconductors} \\ \midrule
Other & $6\%$ & Other, Open\\\bottomrule
\end{tabular}
\vspace{10pt}
\caption{The percentage of Autocast questions in each category, and the subcategories belonging to each category. Autocast questions have fairly even coverage of a wide variety of topics.}
\end{table}

\begin{figure}[t]
    \vspace{5pt}
    \centering
    \includegraphics[width=\textwidth]{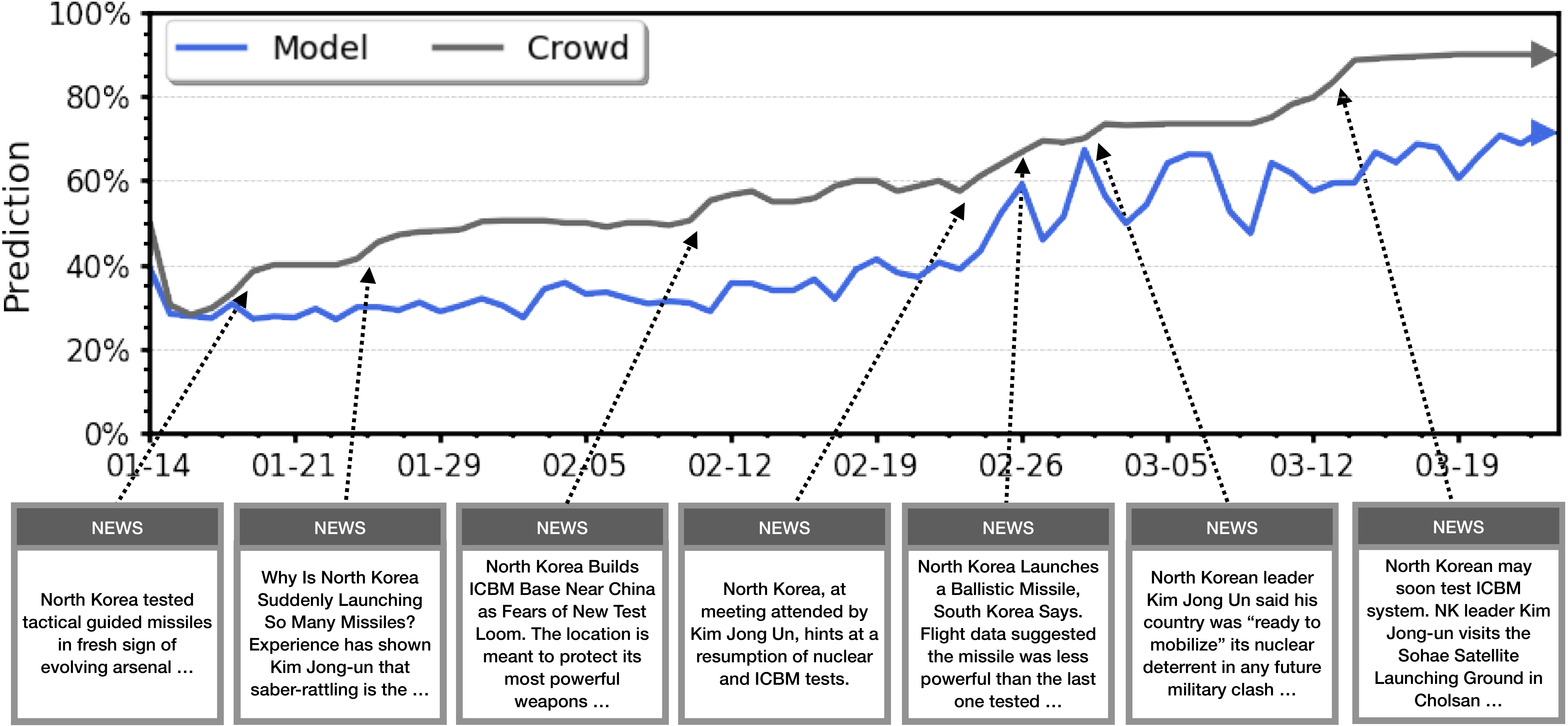}
    \vspace{5pt}
    \caption{The same example from the Autocast dataset shown in Figure \ref{fig:banner}, illustrating how the crowd forecast is influenced by news articles published throughout the prediction period.}
    \label{fig:appendix_splash}
    \vspace{5pt}
\end{figure}

\newpage
\section{X-Risk Sheet}
We provide an analysis of our paper's contribution to reducing existential risk from future AI systems following the framework suggested by \citep{hendrycks2022x}. Individual question responses do not decisively imply relevance or irrelevance to existential risk reduction.

\subsection{Long-Term Impact on Advanced AI Systems}
In this section, please analyze how this work shapes the process that will lead to advanced AI systems and how it steers the process in a safer direction.

\begin{enumerate}[leftmargin=*]
    \item \textbf{Overview.} How is this work intended to reduce existential risks from advanced AI systems?
    
    \textbf{Answer:} This work builds towards improving institutional decision making and systemic safety. In short, this could help resolve matters of fact that influence policies and decisions made by political leaders in an increasingly complex modern world, putting humanity in a better place to deal with the global turbulence and uncertainty created by AI systems when they rapidly reshape society. A fuller motivation for ``ML for Improving Epistemics'' is described in \cite{hendrycks2022x}.
    
    \item \textbf{Direct Effects.} If this work directly reduces existential risks, what are the main hazards, vulnerabilities, or failure modes that it directly affects?
    
    \textbf{Answer:} This directly works against failure modes such as eroded epistemics and hazards such as highly persuasive or manipulative AI systems.
    
    \item \textbf{Diffuse Effects.} If this work reduces existential risks indirectly or diffusely, what are the main contributing factors that it affects?
    
    \textbf{Answer:} This work could lead to improved decision making, epistemics, and collective intelligence. Automated forecasting tools could eventually assist various levels of the sociotechnical hierarchy, including congress and legislatures; government regulatory agencies, industry associations, user associations, etc.; and company management. This lowers the risk of conflict that would accelerate the weaponization of AI, so it diffusely works against weaponized AI failure modes.
    
    \item \textbf{What’s at Stake?} What is a future scenario in which this research direction could prevent the sudden, large-scale loss of life? If not applicable, what is a future scenario in which this research direction be highly beneficial? 
    
    \textbf{Answer:} Advanced automated forecasting better enables political leaders to avoid precarious moments that could spark a large-scale conflict.
    
    \item \textbf{Result Fragility.} Do the findings rest on strong theoretical assumptions; are they not demonstrated using leading-edge tasks or models; or are the findings highly sensitive to hyperparameters? \hfill $\square$
    
    \item \textbf{Problem Difficulty.} Is it implausible that any practical system could ever markedly outperform humans at this task? \hfill $\square$
    
    \item \textbf{Human Unreliability.} Does this approach strongly depend on handcrafted features, expert supervision, or human reliability? \hfill $\square$
    
    \item \textbf{Competitive Pressures.} Does work towards this approach strongly trade off against raw intelligence, other general capabilities, or economic utility? \hfill $\square$
\end{enumerate}

\subsection{Safety-Capabilities Balance}
In this section, please analyze how this work relates to general capabilities and how it affects the balance between safety and hazards from general capabilities.

\begin{enumerate}[resume,leftmargin=*]
    \item \textbf{Overview.} How does this improve safety more than it improves general capabilities?
    
    \textbf{Answer:} While this line of work reduces systemic risk factors and can improve institutional decision making, making AI systems better at forecasting could potentially improve general capabilities. Its relation to general capabilities is currently unclear. In humans, at the extremes, IQ is hardly predictive of forecasting ability, suggesting forecasting of near-term geopolitical events is a specific and not general skill. Likewise, work in this space could focus on engineering better forecasting systems rather than improving general representations, so as to avoid capabilities externalities; this is potentially a more robust strategy for avoiding capabilities externalities. If it turns out that capabilities externalities are difficult to avoid even while simply engineering better forecasting systems, we would suggest that safety researchers stop working on this problem.
    
    \item \textbf{Red Teaming.} What is a way in which this hastens general capabilities or the onset of x-risks?
    
    \textbf{Answer:} Making AI systems better at forecasting could also improve general capabilities or at least the raw power of AI systems. As Yann LeCun reminds us, ``prediction is the essence of intelligence.''
    
    \item \textbf{General Tasks.} Does this work advance progress on tasks that have been previously considered the subject of usual capabilities research? \hfill $\square$
    
    \item \textbf{General Goals.} Does this improve or facilitate research towards general prediction, classification, state estimation, efficiency, scalability, generation, data compression, executing clear instructions, helpfulness, informativeness, reasoning, planning, researching, optimization, (self-)supervised learning, sequential decision making, recursive self-improvement, open-ended goals, models accessing the internet, or similar capabilities? \hfill $\boxtimes$
    
    \item \textbf{Correlation With General Aptitude.} Is the analyzed capability known to be highly predicted by general cognitive ability or educational attainment? \hfill $\square$
    
    \item \textbf{Safety via Capabilities.} Does this advance safety along with, or as a consequence of, advancing other capabilities or the study of AI? \hfill $\boxtimes$
\end{enumerate}

\subsection{Elaborations and Other Considerations}
\begin{enumerate}[resume,leftmargin=*]
    \item \textbf{Other.} What clarifications or uncertainties about this work and x-risk are worth mentioning?
    
    \textbf{Answer:} Regarding Q7, while human forecasters are important for building a training set with rich annotations, the actual human forecasts are unnecessary, as technically only the resolutions are needed. Additionally, the end goal is to create automated forecasting systems that do not depend on human reliability. Eventually, these systems could become much faster and more reliable than human forecasters.
    
    Regarding Q12, this work facilitates research towards general prediction of future events and consequently toward improved planning. However, we expect the kinds of predictions improved by forecasting research to be especially relevant for reducing x-risk. For example, improved institutional decision making surrounding geopolitical events could reduce the risk of global conflicts leading to the weaponization of strong AI.
    
    Regarding Q13, IQ is predictive of forecasting ability in humans, not overwhelmingly so \citep{mellers2015psychology}. Moreover, its correlation is especially weak at extremes. Likewise, forecasting skills for near-term geopolitical events are partly learnable, further suggesting a separation from general cognitive ability.
    
    Regarding Q14, while the relationship between general capabilities and research on forecasting near-term geopolitical events is currently unclear, this research does advance the study of narrow AI systems.

    Finally, we would like to discuss limitations and potential hazards of relying on ML for forecasting near-term geopolitical events.
\begin{enumerate}[leftmargin=*]
    \item Forecasting is best used for refining understanding rather than for anticipating the future more generally. Forecasters are demonstrated to be useful for optimizing probabilities for somewhat likely events (e.g., events with probabilities between, say, 5\% and 95\%). What is more important are tools that unearth important considerations that were implicitly assigned negligible probabilities or wrongly treated by humans as misinformation or worth ignoring. These considerations are often not forecasted and are not thought worth asking; implicitly, such events could the thought to be assigned low probabilities (e.g., say $10^{-7}$), while some people argue that these considerations are more likely than others believe (e.g., say $10^{-1}$). The information value provided from putting ignored considerations on our radar is substantial, in fact, orders of magnitude greater than the information gained by refining probabilities by a few percent. Forecasting competitions are about refining estimates of known unknowns--questions already on our radar--but what is better for risk reduction is confronting unknown unknowns, finding considerations to put on our radar, and reducing \textit{exposure} to inchoate potential risks. For this reason, \cite{hendrycksmlsafety2021} suggest tools that improve brainstorming and suggesting considerations.

    \item Forecasting is not necessarily a suitable tool for addressing tail risks.
    \cite{Taleb2013OnTD} remind us that ``No one has yet
figured out how to design a forecasting tournament to assess
the accuracy of probability judgments that range between .00000001\% and 1\%---and if someone ever did, it is unlikely that anyone would have the patience--or lifespan--to run the forecasting tournament for the necessary stretches of time (requiring us to think not just in terms of decades, centuries and millennia).'' \cite{Taleb2013OnTD} further remind us that it is unjustified to use forecasting tools for revolutions, market crashes, venture capital, or other winner-take-all domains. Furthermore they note that framing questions about tail risks as ``a binary question is dangerous because it masks exponentially escalating tail risks.'' Consequently, ``improving short-run probability judgments'' and ``contingency planning for systemic [tail] risks'' are ``complementary'' and separate \citep{Tetlock2022FalseDA}. Indeed, superforecasters usually anchor in outside view \citep{tetlock2016superforecasting}, which neglects systemic risks. In environments with tail events, it is not how often one is correct that matters but rather how large one's cumulative errors are; current forecasting metrics do not sufficiently penalize forecasters that ignore tail risks nor do they greatly reward prescience about Black Swans.

    \item Forecasting tools could lead to risky behavior. For example, forecasting systems may induce inaction. If forecasts are uncertain, leaders may argue that ``we should not make a decision before we have a reliable forecast'' so we should ``sit tight and assess.'' This is sometimes referred to as the delay fallacy, namely ``if we wait we will know more about X, hence no decision about X should be made now'' \citep{hansson2004fallacies}. However, it is often cheaper to prevent risks or reduce exposure to risks, as ``an existential risk needs to be killed in the egg, when it is still cheap to do so'' \citep{Taleb2020OnSP}. Waiting until all the relevant information arrives is often waiting until it is too late.
    
    Furthermore, humans are known to misinterpret probabilities \citep{Vodrahalli2022UncalibratedMC}. Systems that assign an event 3\% probability may lead decision-makers to assume the event will not happen. Automation bias may mean forecasting systems induce users to have a gain in confidence that is greater than their gain in knowledge. Risk compensation suggests this could result in riskier actions \citep{Hedlund82}. Furthermore, forecasts are often not provided with reverse psychology in mind. However, a forecasting system that forecasts a low risk can lead users to act as though there is no risk and increase risky behavior, which increases systemic risk.

\end{enumerate}

\end{enumerate}

\end{document}